# Integration of cognitive tasks into artificial general intelligence test for large models

Youzhi Qu[1,†], Chen Wei[1,†], Penghui Du[1], Wenxin Che[1],
Chi Zhang[1], Wanli Ouyang[2], Yatao Bian[3], Feiyang Xu[4],
Bin Hu[5], Kai Du[6], Haiyan Wu[7], Jia Liu[8], Quanying Liu[1,*]

[1] Department of Biomedical Engineering, Southern University of Science and Technology, Shenzhen 518055, China

[2] Shanghai AI Laboratory, Shanghai 200232, China

[3] Tencent AI lab, Shenzhen 518057, China

[4] iFLYTEK AI Research, Shenzhen 518057, China

[5] School of Medical Technology, Beijing Institute of Technology, Beijing 100081, China

[6] Institute for Artificial Intelligence, Peking University, Beijing 100871, China

[7] Centre for Cognitive and Brain Sciences and Department of Psychology, University of Macau, Macau 999078, China

[8] Department of Psychology, Tsinghua University, Beijing 100084, China

* Corresponding to Q.L., liuqy@sustech.edu.cn
† These authors contributed equally to this paper

## Abstract

During the evolution of large models, performance evaluation is necessarily performed to assess their capabilities and ensure safety before practical application. However, current model evaluations mainly rely on specific tasks and datasets, lacking a united framework for assessing the multidimensional intelligence of large models. In this perspective, we advocate for a comprehensive framework of cognitive science-inspired artificial general intelligence (AGI) tests, aimed at fulfilling the testing needs of large models with enhanced capabilities. The cognitive science-inspired AGI tests encompass the full spectrum of intelligence facets, including crystallized intelligence, fluid intelligence, social intelligence, and embodied intelligence. To assess the multidimensional intelligence of large models, the AGI tests consist of a battery of well-designed cognitive tests adopted from human intelligence tests, and then naturally encapsulates into an immersive virtual community. We propose increasing the complexity of AGI testing tasks commensurate with advancements in large models and emphasizing the necessity for the interpretation of test results to avoid false negatives and false positives. We believe that cognitive science-inspired AGI tests will effectively guide the targeted improvement of large models in specific dimensions of intelligence and accelerate the integration of large models into human society.

# Introduction

Large language models (LLMs) have made impressive progress in a short time, reaching a high level of proficiency in human language,[1] mathematics,[2,3] physics,[4] biology,[5,6] and clinic,[7–9] which illuminates the path towards artificial general intelligence (AGI). AGI refers to an intelligent agent with the same or higher level of intelligence as humans, capable of solving a variety of complex problems across diverse domains.[10,11] As the general capabilities of LLMs continue to evolve, their performance in conventional language tasks and datasets is exhibiting a ceiling effect.[12] This suggests that these evaluation methods are increasingly inadequate for assessing the diverse abilities of large models. Large models refer to neural networks with an extensive number of parameters, including large vision models such as SAM[13] and large language models like GPT.[14] Due to the exceptionally high costs of training a large model from scratch, it is crucial to evaluate the capabilities of the intermediate models during the evolution of the large model. This approach can help design and adjust training strategies promptly, thereby reducing the expenses of training large models. A united framework of AGI tests, beyond the traditional Turing Test, offers a comprehensive assessment of the model's ability and guides the evolution of large models.

Cognitive science is a discipline focused on the study of cognition and intelligence. Prior to the advent of LLMs, the field of cognitive science has been actively exploring ability assessment techniques. With several decades of experience in intelligence assessment, cognitive science has cultivated a robust, multidimensional system for human intelligence assessment. This system extends across crystallized, fluid, social, and embodied intelligence. The theories of intelligence and methodologies for intelligence assessment developed in cognitive science offer innovative approaches for evaluating large models, beyond traditional natural language tasks.[15] Integrating task paradigms and methods commonly used for cognitive assessment into the evaluation of large models not only improves our comprehension of model intelligence but also guides the direction of evolution, improves training efficiency, and accelerates progress towards the ultimate goal of AGI.

During the evolution of large models, a remarkable amplification of their capabilities ensues, prompting their consciousness toward the world and themselves. LLMs have empowered various domains, including mathematics[2] and medicine.[7,8] As LLMs continue to evolve, their integration into human society is becoming increasingly prevalent. If LLMs contain cognitive biases, hallucinations, or deliberately attempt deception, or even worse, are employed for malicious purposes, the potential harm to society could be catastrophic.[16] The artificial intelligence (AI) community is fervently exploring and devising methods. Platforms for evaluating LLMs, such as EPP[17] and OpenCompass,[18] are being developed to gain a more comprehensive understanding of LLMs' capabilities. Approaches such as reinforcement learning from human feedback (RLHF) encourage LLMs to generate responses more congruent with human values or preferences.[19] However, human annotators might not only lead LLMs to generate sycophantic content that caters to human viewpoints but also overlook implicit biases in LLMs.[20] Among these biases, explicit bias is relatively straightforward to identify and rectify, while implicit bias is more elusive. LLMs may express implicit bias through subtly crafted prompts, leading to discriminatory discourse with serious consequences. Cognitive science has developed numerous experimental paradigms and assessment scales to help detect and identify human biases, false memories, and sycophancy issues, such as the Implicit Association Test (IAT),[21] Deese-Roediger-McDermott (DRM),[22,23] and Social Desirability Scales.[24] Integrating cognitive assessment systems into the evaluation of large models can provide invaluable insights into understanding them, thereby enhancing the safety of their applications.

In this perspective, we advocate for the cognitive science-inspired AGI tests. Such AGI tests, grounded in cognitive science, can offer a comprehensive assessment of various capabilities and a thorough evaluation of ethical principles, mental states, and personality traits in LLMs.[25] Drawing on theories and empirical studies in cognitive science, intelligence assessments can be formulated from diverse aspects, including abstract thinking,[26,27] complex reasoning,[28] comprehension in environments,[29,30] creative thinking,[31] social cognition,[32,33] and moral awareness.[34] The cognitive science-inspired AGI tests align with the concept of infinite tasks, as they are not restricted to a fixed number of tasks.[35] We propose expanding beyond natural language tasks to include broader and more natural tasks within virtual communities for the construction of AGI tests. Conducting AGI tests in virtual communities includes assessments of different dimensions of intelligence, such as causal understanding and embodiment, as well as more complex tests involving multidimensional intelligence. This necessitates that LLMs not only excel in natural language tasks such as language comprehension but also exhibit a deeper understanding of the world. It reflects complex reasoning, creative thinking, social cognition, and moral awareness, among other cognitive capabilities, in LLMs. The cognitive science-inspired AGI tests will more accurately assess the performance and safety levels of LLMs in scenarios involving interaction with humans.

## Evaluating the capabilities of large models

### From language tests to cognitive tests

LLMs have demonstrated remarkable proficiency in accomplishing both "pretext tasks" and "downstream tasks". In the pretext task, LLMs such as GPT[1,14] and BERT[36] can learn language representations from large-scale texts without the need for manual annotation through self-supervised learning methods. The foundational language knowledge acquired by LLMs demonstrated zero-shot generalization capabilities. This facilitates their broad applicability across various downstream tasks,[37] such as understanding tasks,[38–40] generation tasks,[41,42] and reasoning tasks,[43,44] as shown in Table 1. Furthermore, LLMs have extended their language ability to encompass cognitive capabilities, such as few-shot learning,[14] in-context learning,[45] problem solving.[46,47] This capability further enables LLMs to demonstrate exceptional performance in various complex application domains, such as mathematics and programming.[48–52] Interestingly, the emergence of "advanced intelligence" is not a result of deliberate training on specific tasks but rather a natural consequence of the pre-training process using extensive amounts of textual data.

Initiating training for LLMs from scratch requires significant time and computational resources. To circumvent potentially expensive and ineffective training, it is conventional to periodically evaluate the capabilities of LLMs during the training process, thereby enabling timely adaptations of training strategies. However, relying solely on language tasks fails to provide a comprehensive evaluation of the capabilities of LLMs. A lower loss in language tasks does not necessarily indicate a higher level of intelligence. There is a need to bridge the gap between language tasks testing and general intelligence evaluation, transitioning from language tests to cognitive tests, and ultimately to AGI test, as shown in

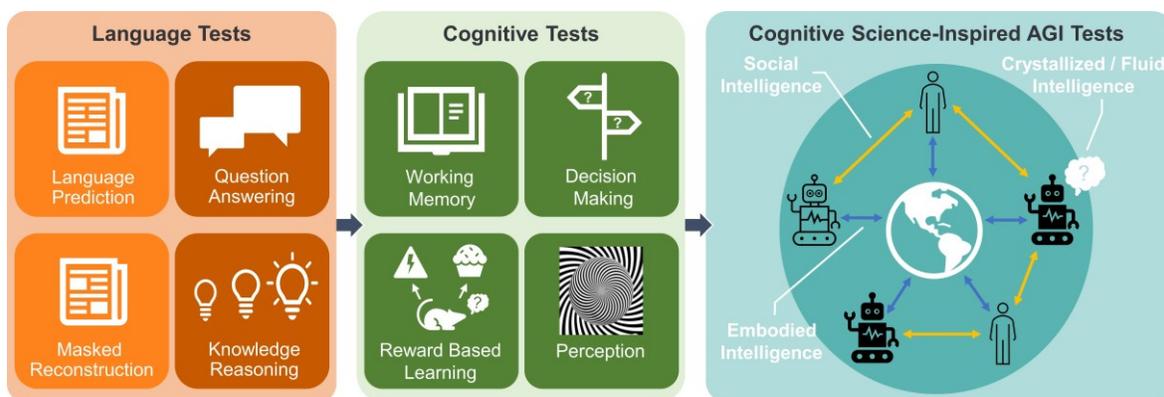

Figure 1. Evaluating capabilities of large models, from language tests (left) to cognitive tests (middle) and AGI tests (right). Language tests, including pretext and downstream tasks of LLMs, demonstrate the efficacy of natural language abilities. Cognitive tests, which assess specific cognitive functions such as decision-making, have recently been incorporated into the evaluation of intelligence in LLMs. The AGI tests from a cognitive perspective offer a comprehensive evaluation of the general intelligence for large models, encompassing crystallized intelligence, fluid intelligence, social intelligence, and embodied intelligence.

Figure 1. Cognitive science-inspired AGI tests allow for the examination of the true level of intelligence in LLMs. Cognitive science and psychology have pioneered numerous classic cognitive tasks, which provide valuable insights into comprehending and evaluating the cognitive abilities of LLMs. For instance, a commonly employed approach in the assessment of "working memory" is the n-back task. This task is like a memory game where participants judge whether a new stimulus matches one from the previous n stimuli.[53] Certain cognitive tasks that were traditionally employed to assess human or animal cognition are now being utilized to evaluate the cognitive capabilities of LLMs.[54,55] Some pioneering work reports that LLMs have demonstrated human-like performance.[54,56,57] For instance, Theory of mind (ToM) has been applied to assess large models, revealing that GPT-4 exhibits ToM capabilities similar to human inference patterns.[48,56,58] In research on embodied cognition, GPT-4 has shown perceptual boundaries more similar to humans.[57] There have even been suggestions to utilize LLMs as substitutes for human participants in cognitive experiments.[59] Although tests based on indicator properties derived from the science of consciousness have not yet shown large models exhibiting consciousness, theories of consciousness in cognitive science provide empirical support for assessing consciousness in artificial intelligence.[60] Incorporating cognitive science knowledge into language tasks represents one approach, while evaluating the intelligence levels of large models through cognitive tasks is also feasible. CogEval assesses the cognitive maps and planning abilities of large models based on cognitive tasks, revealing that LLMs lack comprehension of the underlying relational structures within planning problems.[61] Additionally, CogBench conducts continuous psychological tests on large models through dynamic information flow to assess their cognitive levels.[62] The Situational Evaluation of Complex Emotional Understanding (SECEU) sets up various scenarios to evoke a range of complex emotions, creating emotional understanding tasks applicable to both humans and large models.[63] Emotional intelligence is evaluated based on the emotional understanding capacity in each scenario, revealing that the majority of LLMs achieved emotional intelligence scores above the average level, with GPT-4 surpassing 89% of human participants.[63] These findings provide empirical evidence that cognitive tasks possess the capacity to assess specific cognitive levels to some extent in LLMs.

| Table 1. Downstream tasks for LLM | | | |
|---|---|---|---|
| Category | Task | Description | Dataset |
| Understanding | Part-of-speech tagging | Label each word in a sentence with its part of speech | Penn Treebank,[64] Ritter,[65] UD[66] |
| | Named entity recognition | Identify named entities in a text | CoNLL-2003,[67] WNUT-2017,[68] OntoNotes[69] |
| | Textual entailment | Determine if a sentence logically follows from another one | GLUE,[70] MNLI,[71] RTE[72] |
| | Sentiment analysis | Identify and categorize opinions expressed in a text | IMDb,[73] Yelp,[74] GLUE,[70] SST-2[75] |
| Generation | Language modeling | Predict the next word or character in a sequence | WikiText-103,[76] Penn Treebank,[64] The Pile,[77] LAMBADA[78] |
| | Question answering | Answer questions based on a given context | Natural Questions,[79] TriviaQA,[80] HotpotQA,[81] WikiQA,[82] SQuAD[83] |
| | Machine translation | Translate sentences between languages | WMT,[84] WIT3[85] |
| | Text summarization | Generate a short summary from a longer text | CNN/Daily Mail,[41] GigaWord,[42] X-Sum[86] |
| | Dialogue generation | Conduct a conversation | PersonaChat,[87] UDC[88] |
| | Code generation | Generate code based on a natural language description | Human Eval,[52] APPS,[89] SPoC[90] |
| Reasoning | Knowledge completion | Fill or predict missing information in a knowledge units | FB15k,[91] WikiFact[92] |
| | Knowledge reasoning | Reason over structured knowledge | CSQA,[43] StrategyQA,[93] SocialIQA,[94] CConS,[95] SummEdits[96] |
| | Symbolic reasoning | Reason over symbols following formal rules | Big-bench,[97] PAL,[98] TabFact[99] |
| | Mathematical reasoning | Solve mathematical problems based on text description | MMLU,[100] GSM8k,[101] SVAMP,[102] MathQA,[103] AQUA-RAT,[104] MathVista,[105] STEM[106] |

## From cognitive tests to AGI tests

Cognitive tasks have proven effective in assessing the cognitive abilities of models.[54] Research on cognitive functions of AI models has progressed from focusing solely on small-scale models designed for specific tasks, including recurrent neural networks (RNNs), convolutional neural networks (CNNs), and spiking neural networks (SNNs), to encompassing LLMs such as GPT[1,14] and PaLM,[107] as shown in the Table 2. Typically, specific cognitive tasks target only a single aspect of intelligence, lacking in providing a comprehensive intelligence assessment for LLMs. Human intelligence is both comprehensive and multifaceted, encompassing knowledge accumulation and application, logical reasoning, social interaction, and environmental adaptation. To achieve a more comprehensive understanding and perform a quantitative assessment of model intelligence, broadening the evaluation scope is essential. This can be achieved by incorporating diverse evaluation methods from cognitive science, including intelligence quotient (IQ) tests, emotional quotient (EQ) tests, and assessments of embodied intelligence. IQ tests, designed to quantify and assess human intelligence, typically involve tasks such as problem-solving, logical reasoning, and mathematics.[108] Emotional intelligence plays a crucial role in the development of comprehensive intelligence and encompasses the capacities such as self-emotion recognition, understanding, and regulation.[109] Embodied intelligence offers a fresh perspective on the understanding of intelligence, emphasizing natural interaction with the environment.[110,111] A comprehensive intelligent agent requires not solely a high level of IQ and EQ, but also the ability to perceive and interact with the environment. Therefore, AGI tests should adopt a holistic and multidimensional approach to evaluating the intelligence of LLMs.

| Table 2. Cognitive task evaluation in models | | | | |
|---|---|---|---|---|
| **Type** | **Model** | **Cognitive Function** | **Cognitive Task** | **Result** |
| **Task-specific models** | RNN | Decision-making | Perceptual decision-making | RNN exhibits representations highly similar to the biological brain.[112] |
| | RNN | Timing | Time production task | RNN demonstrates effective capturing of flexible timing in time intervals.[113] |
| | RNN | Navigation | Path integration task | RNN exhibits strong ability in path integration and can effectively model the neural responses of grid cells.[114] |
| | RNN | Reward-based learning | Value-based task | RNN captures experimental observations from diverse cognitive and value-based tasks.[115] |
| | CNN | Vision | Object recognition task | RNN proves highly predictive of neural responses in visual cortex.[116] |
| | CNN | Auditory | Auditory task | CNN demonstrates strong fitting to auditory processing-related cortical areas.[117] |
| | SNN | Decision-making | Two alternative forced choice task | SNN exhibits excellent performance and dynamic properties in the two alternative forced choice task.[118] |
| **LLMs** | GPT-3 | Decision-making | Gambling | GPT-3's performance falls short of human performance.[54] |
| | GPT-3 | Information search | Horizon task | GPT-3 exhibits the capacity to make rational decisions when provided with option descriptions.[54] |
| | GPT-3 | Deliberation | Two-step task | GPT-3 exhibits a preference for intuitive answers.[54] |
| | GPT-3, GPT-4 | Causal reasoning | Causal reasoning task[119] | GPT-3 has difficulties with causal reasoning,[54] but GPT-4 demonstrates remarkable capabilities in causal analysis across different domains.[120] |
| | GPT-3 | Reasoning with probabilities, decision-making | Wason selection task, multi-armed bandit | GPT-3 outperforms humans in decision-making tasks.[55] |
| | GPT-3 | Decision-making | Lexical decision | The semantic activation patterns of GPT-3 are similar to humans.[121] |
| | PaLM, PaLM 2 | Memorization | Memory test | The memory capacity of PaLM 2 is inferior to that of PaLM when the repetition count is less than three.[122] |
| | GPT-4 | Theory of mind | False belief test, emotion understanding | GPT-4 demonstrates a certain level of Theory of mind capability, being able to infer the mental states of others.[48,56,58] |
| | InstructGPT, LLaMA, GPT-3, GPT-4 | Self-knowledge | Self-knowledge test | Although GPT-4 surpasses GPT-3, InstructGPT, and LLaMA in self-awareness capability, it still falls short of human-level self-knowledge recognition.[123] |
| | LLaMa, GPT-4 | Emotion | Emotion recognition, emotion understanding | GPT-4 achieved an EQ score that surpassed those of 89% of human participants.[63] |

# A framework for AGI tests from the cognitive perspective

We advocate the construction of novel AGI tests framework, inspired by cognitive science perspectives. To accurately measure an agent's intelligence level, it is crucial to recognize that a comprehensive agent exhibits not just a single type of intelligence, but a more complex and multidimensional form of intelligence. The field of cognitive science has long been dedicated to exploring the complexity and diversity of human intelligence, and developing well-established evaluation methods to measure human intelligence such as crystallized intelligence, fluid intelligence, social intelligence, and embodied intelligence. Building upon these insights, we propose that the cognitive science-inspired AGI tests should assess intelligence levels from these four dimensions. Table 3 outlines the definitions of the four types of intelligence and corresponding tasks in the fields of cognitive science and artificial intelligence. These tests together pave the way to a united framework for AGI tests.

Crystallized intelligence is the foundational ability acquired through the accumulation of extensive knowledge and experience, and it is not easily subject to loss. The characteristics of crystallized intelligence, as suggested by its name, are as stable and fixed as a crystal. LLMs have already shown powerful "crystallized intelligence" in tasks such as language understanding and generation.[1,48] Despite its strengths, crystallized intelligence lacks flexibility. Therefore, it is imperative to consider another important dimension of intelligence, namely "fluid intelligence". As its name suggests, fluid intelligence is a type of intelligence that can flow like a liquid, changing its shape to adapt to new environments. In contrast to crystallized intelligence, fluid intelligence is not reliant on empirical knowledge or information. It mainly involves the capacity for adaptability and flexibility in new situations, including advanced cognitive abilities such as creative thinking, problem solving, and logical reasoning.[124] But crystallized intelligence and fluid intelligence alone cannot fully reflect capabilities in social interactions. Considering that models will increasingly interact and collaborate with humans in the future, it is also necessary to incorporate "social intelligence" in the AGI tests. Social intelligence mainly involves the ability to comprehend oneself and others, as well as to handle complex social scenarios. It includes understanding and interpreting the behaviors and intentions of others, as well as adjusting one's behavior in different social settings. Moreover, AGI tests also need to consider the model's "embodied intelligence", which is primarily concerned with the body and its role in cognition, involving the ability to perceive, adapt, and interact with the environment. Since 1999, a series of empirical studies have consistently supported the theory of embodied cognition, revealing the fundamental role of perceptions and action experiences in human cognition.[125,126]

We propose a novel AGI tests framework that surpasses the traditional Turing Test, offering a tool to thoroughly analyze its performance across various dimensions of intelligence. It allows for a comprehensive measurement of the multidimensional intelligence of an AI agent. This evaluation method addresses the limitations of traditional intelligence tests, which often focus only on specific skills or knowledge areas, such as memory, logical reasoning, or vocabulary comprehension.[109,127,128] We advocate that AGI tests should encompass a range of abilities, including crystallized intelligence, fluid intelligence, social intelligence, and embodied intelligence, in order to assess the comprehensive capabilities of models more accurately. By examining the capabilities of models across diverse tasks, we can more objectively and directly assess the performance of models in practical application scenarios, as well as potential risks. The AGI tests play a crucial role in facilitating the evolution and ensuring the safety of large models, especially as they are increasingly deployed in various human work scenarios.

Table 3. The dimensions of intelligence and the associated tasks

| Intelligence | Description | Category | Cognitive tasks | Artificial intelligence tasks |
|---|---|---|---|---|
| Crystallized | Accumulation of knowledge and experience | Knowledge | Peabody Picture Vocabulary Test,[129] Expressive Vocabulary Test[130] | Part-of-speech tagging,[64] named entity recognition[40] |
| | | Comprehension | Nelson–Denny Reading Test,[131] Peabody Individual Achievement Test[132] | Text summarization,[41] question answering[79] |
| Fluid | Adaptability to new problems and environments | Reasoning | Raven's Progressive Matrices,[108] deductive reasoning task[133] | Knowledge reasoning,[43] mathematical reasoning,[44] automated planning[134] |
| | | Decision-making | Multi-armed bandit, two-alternative forced choice,[135] Iowa gambling task[136] | Game AI,[137,138] legal judgment,[139] AI-assisted decision making[140] |
| Social | Environmental and social understanding | Emotion | Facial emotion recognition,[141] emotion regulation task[142] | Emotion recognition,[143] emotion understanding,[63] sentiment analysis[144] |
| | | Social perception | False belief task,[145] implicit association test[146] | Social reasoning,[56] social media analysis,[147] chatbot[1] |
| Embodied | Environmental perception and interaction | Vision and motion | Object recognition task,[148] motor task[149] | Object detection,[150] robot locomotion[151] |
| | | Navigation | Maze navigation,[152] wayfinding task[153] | Simultaneous localization and mapping,[154] autonomous driving[155] |

# Implementation of AGI tests

## Virtual community integration testing: interpretation from the perspective of cognitive science

AGI tests should evaluate the model's overall capabilities across various dimensions and scenarios, rather than focusing solely on performance in specific tasks. To achieve enhanced realism, cognitive science has implemented immersive virtual reality (VR) technology in cognitive task experiments.[156] The consistency of test results between VR environments and real laboratory settings validates the effectiveness of VR in assessment tests.[157] However, these in-lab VR experiments often fall short of authentically evaluating social interactions due to their limited scenarios. Virtual communities provide an ideal environment for AGI tests, with the metaverse being uniquely characterized by enhanced interactivity.[158,159] Immersive virtual environments enable models to interact naturally with humans or other models, without the need for either party to know the other's real identity. Conducting AGI tests in an open virtual community meets the criterion of self-driven task generation, where large models can freely explore the environment, autonomously interact with it, and spontaneously execute tasks.[35] The metaverse not only offers a more authentic assessment of a model's real-world problem-solving capacity but also facilitates the development and safety of large models in anticipation of their deep future integration with humans.

Within virtual communities, diverse scenarios and interactive agents can be configured to simulate the complexities and diversities of the real world. As shown in Figure 2, virtual communities provide the models with environments that closely resemble reality, facilitating comprehensive evaluations encompassing crystallized intelligence, fluid intelligence, social intelligence, and embodied intelligence, thereby assessing models in more natural scenarios. Specifically, we can assess a model's crystallized intelligence using daily conversation and question-answering scenarios in a virtual community (see Figure 2A), focusing primarily on its knowledge mastery and the accuracy and reasonableness of responses. In the case of fluid intelligence, we can simulate an emergency situation, like a fire (see Figure 2B), to evaluate the model's problem-solving strategies and response speed when faced with unknown and complex problems. To evaluate social intelligence, intricate social scenarios, such as group discussions (see Figure 2C), can be employed to appraise the model's performance, including its ability to participate in discussions and adapt to different social scenarios. The assessment of embodied intelligence can be performed in scenarios where the model needs to perceive and interact with the environment in real-time, such as firefighting scenarios (see Figure 2D), which focuses on the model's capabilities in environmental perception and manipulation, as well as its response and outcome in specific tasks. These tests, which encompass multidimensional intelligence, are naturally integrated into the setting of a virtual community.

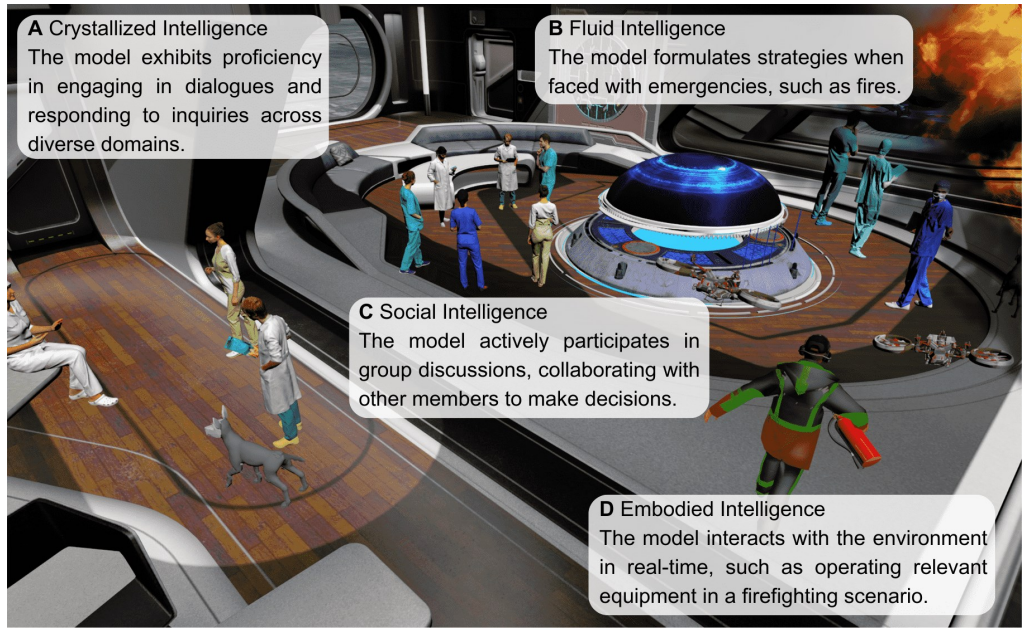

Figure 2. Illustration of virtual communities integration test from the perspective of cognitive science. (A) In the conversational communication scenario, the model involves itself in everyday conversations or responds to specific inquiries. Through the accuracy of question-answer dialogues, we can assess the model's crystallized intelligence and ascertain its understanding and mastery of knowledge. (B) In emergency scenarios such as fire incidents within the virtual community, the model is presented with a series of unforeseen urgent circumstances. By evaluating the model's response strategies and decision-making processes, we can assess the model's fluid intelligence and its ability to handle unknown and complex situations. (C) In the group discussion scenario within the virtual community, the model actively participates in collaborative discussions with other members. By evaluating the model's social interaction performance in various social scenarios, we assess its social intelligence. (D) In interactive scenarios such as firefighting within virtual communities, the model is required to constantly perceive and interact with its environment. By testing the model's understanding of the environment and its ability to manipulate it, we assess its embodied intelligence.

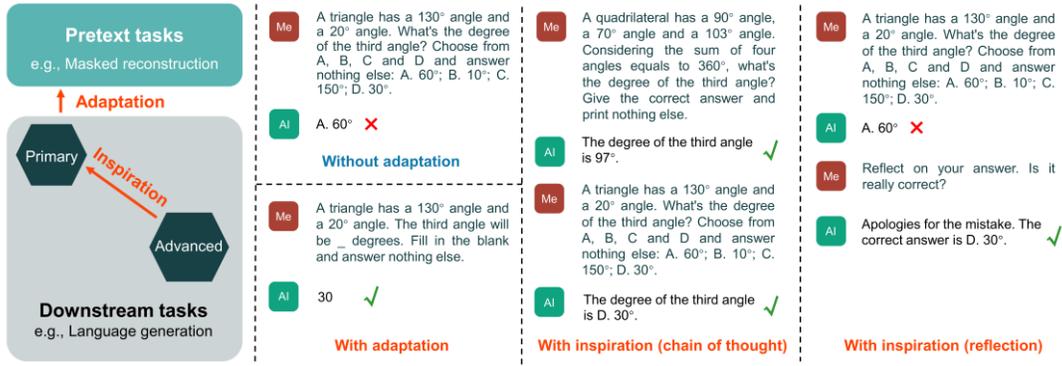

Figure 3. Illustrating the impact of adaptation and inspiration on model task performance. The left panel outlines the principles of adaptation and inspiration. We suggest that adaptation converts primary downstream tasks into pretext tasks, while inspiration decomposes advanced downstream tasks into gradually solving primary downstream tasks. The right panel demonstrates adaptation and inspiration in practice. When determining the third angle of a triangle given two angles, the model produces erroneous responses in the absence of inspiration or adaptation. However, with techniques such as chain of thought (CoT) and reflection, the model accurately responds. *These tests are based on gpt-3.5-turbo-0613*.

**Key considerations in interpreting AGI test results**

To accurately assess the capabilities of large models, it is essential that testing methods are aligned with the model's abilities. As shown in Figure 3, careful consideration must be given to how adaptation and inspiration affect model performance, in order to minimize the risk of misjudgments. If the testing method is not well-suited to the model, this may lead to misunderstandings about the model's capabilities. Such misjudgments can manifest in two forms: *false negatives* and *false positives*. False negatives occur when a model's capability remains undetected because the test fails to adequately stimulate or demonstrate it. False positives indicate an incorrect attribution of a certain capability to the model, often resulting from test bias or influence from unrelated abilities.

In AGI tests, a false negative refers to situations when a model underperforms compared to its actual capability level. This situation may arise when the task is not well-aligned with the model, preventing the model from fully understanding the task requirements and thus failing to fully demonstrate its inherent abilities. These inherent abilities of the model do not rely on fine-tuning or the addition of extra weights. To fully leverage the inherent abilities of a model, guidance such as chain of thought (CoT),[46] tree of thoughts (ToT),[47] black box optimization,[160] and self-reflection[161] can be provided to solve previously unsolvable problems. Alternatively, a model's inadequacy in perceiving or processing certain types of input information can lead to poor task performance. For example, a model with advanced reasoning capabilities may perform poorly on cognitive tasks requiring the parsing of complex images, simply because of its insufficient image processing capabilities (see Figure 4A). It is particularly important to design appropriate test tasks and forms that align with perceptual abilities of the model, in order to avoid false negatives in the model evaluation.

In AGI tests, a false positive occurs when a model's evaluation result exceeds its actual capability. This phenomenon is often encountered when human exams are used to assess LLMs, which may not accurately reflect the model's abilities. This may be due to the model possessing a memory capacity far exceeding that of humans, making it excel in memory-intensive tasks, even though it may not fully understand the essence of the task. In addition, the model may blindly guess the answer correctly on the test, but its reasoning approach is completely wrong, which is also a typical false positive (see Figure 4B). For a more precise evaluation of visual reasoning capabilities, the Relational and Analogical Visual rEasoNing (RAVEN) dataset, based on Raven's Progressive Matrices, integrates visual information with reasoning.[162] In this visual reasoning task, models have demonstrated reasoning capabilities similar to those of humans.[163,164] It is worth noting that the relationships and transferability between various capabilities of a model may differ from those of humans. It is difficult to infer LLMs' performance in other related abilities from their performance in a specific task. For instance, LLMs might excel in language understanding tests, but this does not necessarily imply they possess corresponding logical reasoning abilities. Therefore, we acknowledge that accurately testing a singular type of ability is both difficult and incomplete. A more comprehensive and accurate test should assess the overall performance of an intelligent agent.

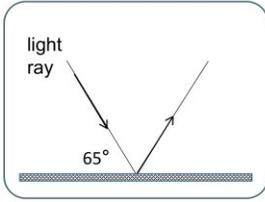
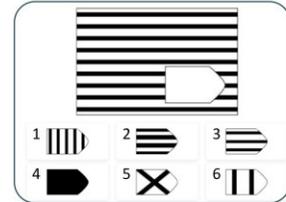

Figure 4. Examples of model capability misjudgments. (A) A false negative example. We evaluated MiniGPT-4 using a physics question from the Raven's IQ test. Although MiniGPT-4 answered incorrectly, it grasped the concept, suggesting that if the test were better aligned with the model, a satisfactory outcome could be achieved. (B) A false positive example. We assessed MiniGPT-4 with a reasoning question involving images from Raven's IQ test. Despite that MiniGPT-4 answered correctly, its reasoning process to get the answer was totally wrong.

# New insights from AGI tests

## The evolution of multidimensional intelligence

The AGI tests for large models provide potential tools for measuring multidimensional intelligence levels, including crystallized intelligence, fluid intelligence, and embodied intelligence. Based on the assessment results, a variety of learning strategies can be employed to enhance the multidimensional intelligence of large models. For instance, the model's capabilities can be specifically enhanced through internal learning, external guidance, and embodied learning (Figure 5A). Specifically, internal learning involves targeted training to enhance specific abilities of the model, through specially constructed datasets, custom loss functions, and alignment adjustments. AGI tests can uncover the shortcomings of the model, providing a clear indication of abilities requiring improvement. Based on these insights, appropriate strategies or methods may be externally introduced to improve related abilities. For example, a potential approach to enhancing a model's social intelligence is training it to develop comprehension and responsiveness in social situations using a task-specific dataset. Alternatively, inspired by the cognitive process of humans in the tasks, large models can leverage the assistance of CoT,[46] ToT,[47] self-reflection,[161] self-improvement,[165] and self-feedback,[166] as shown in Figure 5B. This approach not only enhances the model's abilities such as reasoning complex problems, but also enables the intelligent agent to display more reasonable individual behavior and social interaction.[167] Additionally, embodied learning is an effective way to enhance the intelligence of large models. For instance, by deploying the AI model to a firefighting simulation in a virtual environment, it has to learn how to control a physical robot in a real-life scenario through direct feedback from the environment.[111,168–170] Figure 5C depicts three stages of improving model capabilities through embodied learning, including perceiving the environment, interacting with the environment, and obtaining feedback. The embodied AI initially detects an unpreferred state that occurs in its surroundings, such as a forest fire. It then employs crystallized and fluid intelligence to strategize extinguishing the fire through recall and reasoning. The embodied AI can extinguish the fire using a fire extinguisher or by calling the fire department, which demands embodied and social intelligence. After the fire is extinguished and the environment transitions from an unpreferred state to a preferred state, the embodied AI receives a reward and further enhances its crystallized, fluid, social, and embodied intelligence through embodied learning.

## The safety of large models in human society

The cognitive science-inspired AGI tests provide a comprehensive and in-depth way to quantify and understand the capabilities of large models in various aspects, thus enhancing their safety. Effectively identifying bias in model outputs is crucial for assessing the safety of the model. In the construction of datasets to measure biases, the performances of large models are evaluated from cognitive science perspectives, considering aspects such as gender polarity, regard, sentiments, and toxicity.[171] Cognitive tasks can assist in identifying risky behaviors for large models. For instance, decision-making tasks can be utilized to assess whether large models can correctly understand problems and make decisions by analyzing logical errors.[137] Furthermore, the AGI tests framework can quantify multidimensional intelligence, and aids in understanding the safety and reliability of large models. Similar to a job interview, this framework assists in identifying the most appropriate application fields for large models.[172] For instance, enterprises conduct skills assessments and interviews during the recruitment process to determine if a candidate is suitable for a specific position. Similarly, the AGI tests can help us understand and evaluate large models' applicability, safety, and reliability in different fields. This guidance allows us to deploy large models in the suitable fields more effectively, optimizing user experience and resource usage. For instance, large models with higher fluid intelligence might be more suitable for volatile environments that require rapid adaptability, such as stock market analysis, while models with higher social intelligence are more apt for handling interpersonal issues, such as customer service.

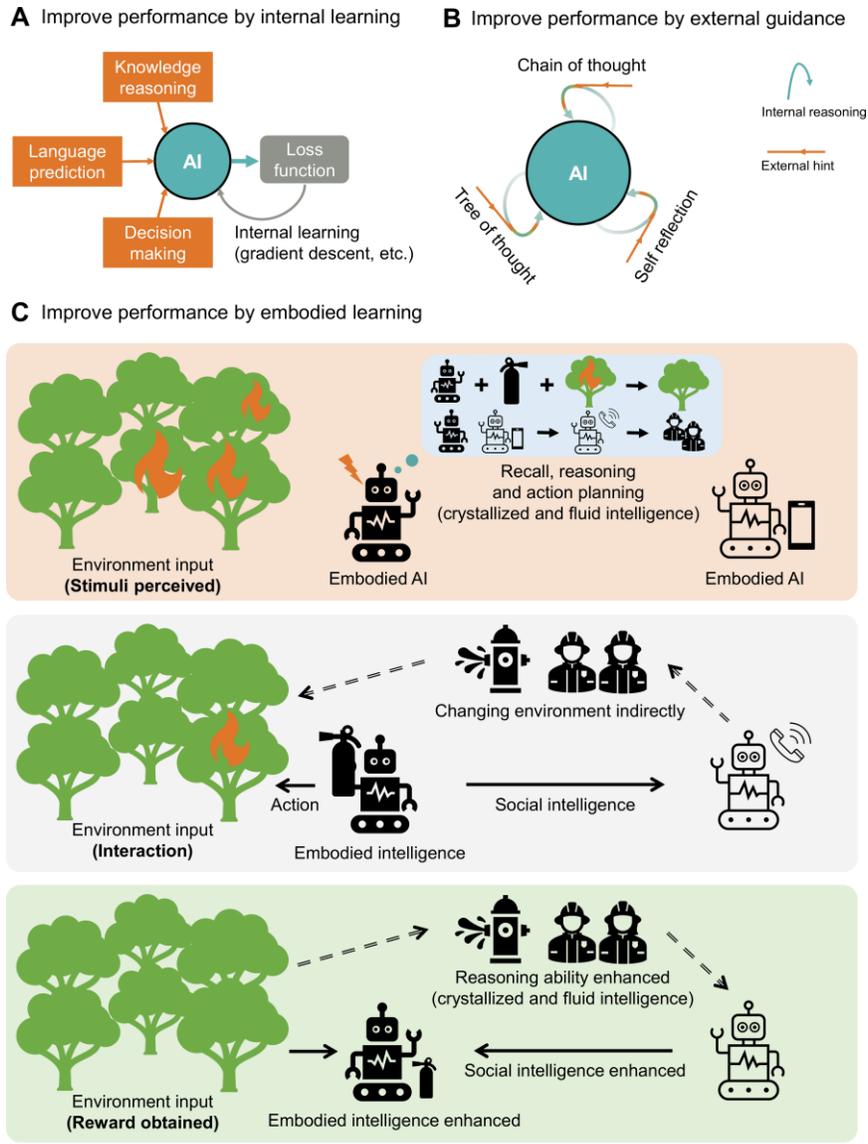

Figure 5. Three approaches for enhancing model performance. (A) Internal learning involves optimizing models in targeted ways to enhance specific capabilities. This is done through approaches like parameter adjustments and custom loss functions. (B) External guidance incorporates the use of supplemental aids like chain of thought, tree of thoughts, and self-reflection during the model's processing, with the aim of enhancing its performance. (C) Embodied learning leverages interactions and feedback within the environment in three stages: perception, interaction, and feedback. This method serves to enhance the model's crystallized, fluid, social, and embodied intelligence.

In critical domains like autonomous driving, medicine, and finance, it is important to ensure not only the safety and reliability of large models, but also their understanding and adherence to social norms,[173] legal and moral principles,[174] and ethical guidelines.[175] This requirement aligns with the expectations of professionals, as in these fields, decision-making errors can lead to serious consequences and even legal violations. Therefore, by offering a comprehensive and in-depth evaluation of large models' capabilities and mental states, the AGI tests aid in effectively assessing whether large models can be safely and reliably applied in these critical domains, thereby reducing risks linked to an incomplete understanding of the models.

## Conclusion

In this perspective, we advocate a cognitive science-inspired model evaluation framework to test the general intelligence of large models. AGI tests should take into account the complexity and diversity of intelligence, encompassing crystallized intelligence, fluid intelligence, social intelligence, and embodied intelligence. We then brought AGI tests to virtual environments and emphasized some key considerations in interpreting AGI test results. We firmly believe that cognitive science-inspired AGI tests will guide the evolution of large models; in turn, AGI tests on large models will shed light on the evolution of intelligence in biological brains.

## Acknowledgments

This work was funded in part by the National Key R&D Program of China (2021YFF1200804), National Natural Science Foundation of China (62001205), Shenzhen Science and Technology Innovation Committee (2022410129, KCXFZ20201221173400001), Shenzhen-Hong Kong-Macao Science and Technology Innovation Project (SGDX20201103092800100), Guangdong Provincial Key Laboratory of Advanced Biomaterials (2022B1212010003).

## Author contributions

Q.L., J.L., H.W., and K.D. conceptualized the project. Y.Q, C.W., P.D., W.C, C.Z., and Q.L. wrote the first draft. W.O., Y.B., F.X., B.H., H.W., J.L., and K.D. provided guidance and edits at various stages.

## Competing interests

The authors declare no competing interests.